\documentclass{article}

\usepackage{arxiv}

\usepackage[utf8]{inputenc} 
\usepackage[T1]{fontenc}    
\usepackage{hyperref}       
\usepackage{url}            
\usepackage{booktabs}       
\usepackage{amsfonts}       
\usepackage{nicefrac}       
\usepackage{microtype}      
\usepackage{lipsum}

\usepackage{comment}
\usepackage{bm}
\usepackage{amsmath}
\usepackage{amssymb}
\usepackage{amsfonts}
\usepackage{graphicx}
\def\vec#1{\mathbf{#1}}
\usepackage{algorithm}
\usepackage{algorithmic}
\newcommand{\argmax}{\mathop{\rm arg~max}\limits}

\title{End-to-End Learning of Deep Kernel Acquisition Functions for Bayesian Optimization}

\author{
  Tomoharu Iwata\\
  NTT Communication Science Laboratories\\
}
\date{}

\begin{document}
\maketitle

\begin{abstract}
For Bayesian optimization (BO) on high-dimensional data with complex structure, neural network-based kernels for Gaussian processes (GPs) have been used to learn flexible surrogate functions by the high representation power of deep learning. However, existing methods train neural networks by maximizing the marginal likelihood, which do not directly improve the BO performance. In this paper, we propose a meta-learning method for BO with neural network-based kernels that minimizes the expected gap between the true optimum value and the best value found by BO. We model a policy, which takes the current evaluated data points as input and outputs the next data point to be evaluated, by a neural network, where neural network-based kernels, GPs, and mutual information-based acquisition functions are used as its layers. With our model, the neural network-based kernel is trained to be appropriate for the acquisition function by backpropagating the gap through the acquisition function and GP. Our model is trained by a reinforcement learning framework from multiple tasks. Since the neural network is shared across different tasks, we can gather knowledge on BO from multiple training tasks, and use the knowledge for unseen test tasks. In experiments using three text document datasets, we demonstrate that the proposed method achieves better BO performance than the existing methods.
\end{abstract}

\section{Introduction}

Bayesian optimization (BO) is an approach for
the global optimization of black-box functions
that are expensive to evaluate~\cite{pelikan1999boa,brochu2010tutorial,shahriari2016taking}.
BO has been successfully used for a wide variety of applications,
such as 
computer vision~\cite{denil2012learning},
recommendations~\cite{gonzalez2017preferential,galuzzi2019bayesian},
chemical design~\cite{griffiths2017constrained},
material science~\cite{seko2015prediction,ueno2016combo},
probabilistic programs~\cite{rainforth2016bayesian},
and the automatic selection of machine learning algorithms~\cite{snoek2012practical,bergstra2013hyperopt,thornton2013auto,kotthoff2017auto,klein2017fast}.
Gaussian processes (GPs)~\cite{rasmussen2006gaussian}
are commonly used as surrogates to model a distribution over target functions.
The next data point to be evaluated is selected by an acquisition function
calculated using GPs in such for finding optimal points with fewer
target function evaluations.

With GPs, it is important to use appropriate kernel functions.
Neural networks have been used
as components of kernel functions~\cite{wilson2011gaussian,huang2015scalable,calandra2016manifold,wilson2016deep,wilson2016stochastic},
which enable us to learn flexible kernel functions for
high-dimensional data with complex structure
by the high representation power of deep learning.
For example, neural network-based kernels are obtained by
transforming the input of RBF kernels by neural networks.
However, many training data are required
for training neural networks.
Therefore, the use of neural networks in BO is limited,
since obtaining a large amount of training data is expensive.

To learn expressive kernels
with a small amount of training data in a test task,
transfer learning and meta-learning methods have been
proposed~\cite{yu2005learning,bonilla2008multi,wei2017source,iwata2019efficient,harrison2018meta,tossou2019adaptive,iwata2020few,patacchiola2020bayesian}.
With such methods,
the knowledge learned from training tasks
is transferred to test tasks,
where the training tasks are related to but different from test tasks.
These existing methods train neural network-based GPs
by maximizing the marginal likelihood.
Therefore, although they improve the performance of estimating
target functions, they do not necessarily improve the BO performance
since it is not directly optimized,
and the acquisition function to be used is not considered for training.

In this paper, we propose a meta-learning method
for training neural network-based kernels
by directly maximizing the BO performance
with the consideration of the acquisition function to be used.
We model a policy, which takes the current evaluated data points as input
and outputs the next data point to be evaluated,
a neural network,
where neural network-based kernels, GPs, and mutual information-based
acquisition functions (MI)~\cite{contal2014gaussian} are used as its layers.
By incorporating the MI into the policy network,
the neural networks in the kernel function are trained
such that the BO performance is improved when MI is used.
We call our model the deep kernel acquisition function.
Since the predictive mean and variance given the evaluated data points
are calculated in a closed form by GP, and
the MI is also calculated in a closed form,
we can backpropagate the gap through the GP and MI
to update the neural network parameters.
We can use other acquisition functions,
such as expected improvement~\cite{mockus1978application,jones2001taxonomy}
and upper confidence bound~\cite{auer2002using,srinivas2010gaussian},
instead of MI if they are differentiable.

\begin{figure}[t!]
  \centering
  \includegraphics[width=35em]{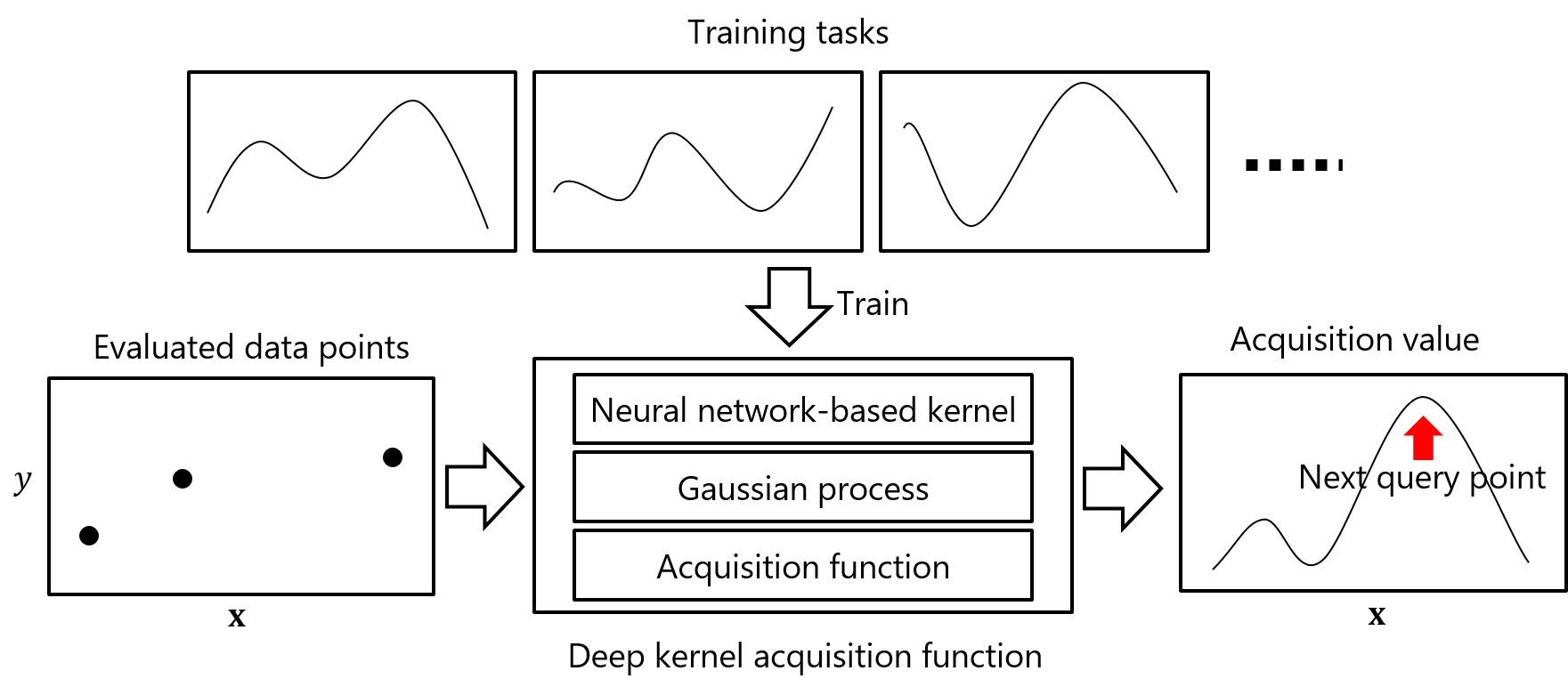}
  \caption{Framework of our proposed method. We are given multiple tasks for training, where feature vectors and their response values are known. Training tasks are used for training our model (deep kernel acquisition function), which consists of neural network-based kernels, Gaussian processes, and mutual information-based acquisition functions. Our model takes a set of evaluated data points as input, and outputs acquisition values. Using the acquisition values, a data point is selected to be evaluated next. For each training epoch, first, a task is randomly selected from the training tasks. Second, we run Bayesian optimization using our model on the selected task and evaluate the gap. Third, the parameters of the neural network-based kernels are updated by minimizing the gap using reinforcement learning, where a loss is backpropagated through the acquisition function and the Gaussian process.}
  \label{fig:method}  
\end{figure}

We formulate BO as a Markov decision process,
where the current evaluated data points are a state,
the next data point to be evaluated is an action,
and the BO performance is a reward.
For evaluating the BO performance,
we use the gap~\cite{wang2016bayesian,berkenkamp2019no}
which is the difference between the true optimum value
and the best value found.
The proposed method trains the policy network
by minimizing the expected gap using
the policy gradient method~\cite{sutton1999policy}
with a reinforcement learning framework.
The expected gap is calculated by
running BO procedures on randomly sampled training tasks
for each training epoch using
an episodic training framework~\cite{ravi2016optimization,santoro2016meta,snell2017prototypical,finn2017model,li2019episodic}.
Since the policy network is shared across different tasks,
we can gain useful knowledge for BO from a wide variety of training tasks,
and use it for unseen test tasks that
are different from training tasks.
Figure~\ref{fig:method} illustrates the training framework of the proposed method.

\section{Related work}
\label{sec:related}

BO has been formulated as reinforcement learning problems~\cite{volpp2019meta,vishnu2019meta}.
MetaBO~\cite{volpp2019meta} models an acquisition function
by a neural network that takes the outputs of Gaussian processes,
which are the mean and variance of the response value at a query point,
as input. The kernel parameters of GPs are estimated
by maximizing the marginal likelihood to fit the training data,
and then the neural network parameters are estimated by reinforcement learning.
On the other hand, the kernel parameters in the proposed method
are estimated to improve the BO performance,
which enables us to estimate the mean and variance
of the response value that are appropriate for BO.
Vishunu et al.~\cite{vishnu2019meta} uses recurrent neural networks
to model a policy that takes the sequence of evaluated data points as input.
In their model, they do not use GPs,
although GPs have been successfully used for a wide variety of BO problems.
Other works
\cite{volpp2019meta,vishnu2019meta}
completely depend on neural networks
for modeling acquisition functions,
which require much training data.
In contrast to all of the existing methods,
we can exploit the knowledge on acquisition functions
in the BO literature with the proposed method using
a mutual information-based acquisition function
as a layer in the neural network.
It would help to improve the performance
especially with a small number of training data
since we do not need to train the neural network from scratch.
Although many meta-learning methods have been proposed~\cite{schmidhuber:1987:srl,bengio1991learning,finn2017model,vinyals2016matching,snell2017prototypical,garnelo2018conditional}
including GP-based methods~\cite{harrison2018meta,tossou2019adaptive,fortuin2019deep,iwata2020few},
they are designed for supervised learning,
but not for BO.

\section{Proposed method}
\label{sec:proposed}

\subsection{Problem formulation}

Suppose that we are given $S$ datasets on
BO tasks
$\{\{(\vec{x}_{sn},y_{sn})\}_{n=1}^{N_{s}}\}_{s=1}^{S}$ in the training phase,
where
$\vec{x}_{sn}\in\mathcal{X}$
is the $n$th feature vector in the $s$th task,
$y_{sn}=f_{s}(\vec{x}_{sn})$ is its scalar response value
by task-specific black-box function $f_{s}(\cdot)$,
and $N_{s}$ is the number of examples in the $s$th dataset.
We assume that feature spaces $\mathcal{X}$ are identical across tasks.
The number of data points can be different across tasks.
In the test phase,
we are given a set of feature vectors for a test task
$\mathcal{D}_{*}=\{\vec{x}_{*n}\}_{n=1}^{N_{*}}$,
where
$\vec{x}_{*n}\in\mathcal{X}$ is the $n$th feature vector in the test task.
Although the test task is related to some of the training tasks,
it is different.
Our goal is to find point $\vec{x}\in\mathcal{D}_{*}$
that has higher response value $f_{*}(\vec{x})$ with fewer queries,
where $f_{*}(\cdot)$ is the target function of the test task we want to maximize.
For simplicity, we assume that a set of feature vectors
for each task is given. When no set is given,
the proposed method is applicable
by generating a set of feature vectors.

\subsection{Gap minimization}

For each timestep in the $s$th task,
BO selects
a data point in $\mathcal{D}_{s}=\{\vec{x}_{sn}\}_{n=1}^{N_{s}}$
to evaluate next using policy function $a(\cdot)$:
\begin{align}
  n_{s,t+1}=\argmax_{n\notin\mathcal{N}_{st}}a(\vec{x}_{sn};\mathcal{N}_{st},\bm{\Psi}),
  \label{eq:n}
\end{align}
where $n_{st}\in\{1,\cdots,N_{s}\}$
is the data point index of the query at the $t$th timestep in the $s$th task,
$\mathcal{N}_{st}\subseteq\{1,\cdots,N_{s}\}$
is the set of evaluated data point indices until the $t$th timestep,
$n\notin\mathcal{N}_{st}$ represents a data point index
that is not included in evaluated data points $\mathcal{N}_{st}$,
and $\bm{\Psi}$ is the parameters of the policy function.
When $\mathcal{N}_{st}$ is the evaluated data points, we assume that
their feature vectors and response values 
$\{(\vec{x}_{sn},y_{sn})\}_{n\in\mathcal{N}_{st}}$ are observed,
and omit the dependency of $\{(\vec{x}_{sn},y_{sn})\}_{n\in\mathcal{N}_{st}}$ in the equations.

For the evaluation measurement of BO, the following expected cumulative gap is used,
\begin{align}
  R = \mathbb{E}_{s}\left[\sum_{t=1}^{T}p(\mathcal{N}_{st})r_{s}(\mathcal{N}_{st})
    \right],
  \label{eq:R}
\end{align}
where $\mathbb{E}_{s}[\cdot]$ denotes the expectation over tasks,
$T$ is the number of data points to be evaluated,
$p(\mathcal{N}_{st})$ is the probability that the evaluated data points are $\mathcal{N}_{st}$ at the $t$th timestep,
and
\begin{align}
  r_{s}(\mathcal{N}_{st})=\max_{n\in\{1,\cdots,N_{s}\}}y_{sn}-\max_{n\in\mathcal{N}_{st}}y_{sn},
  \label{eq:r}
\end{align}
is the difference between true maximum value
$\max_{n\in\{1,\cdots,N_{s}\}}y_{sn}$
and the maximum value in the evaluated data points until the $t$th timesteps 
$\max_{n\in\mathcal{N}_{st}}y_{sn}$
in the $s$th task.
Since we are given the set of feature vectors and their response values
$\{(\vec{x}_{sn},y_{sn})\}_{n=1}^{N_{s}}$ for the training tasks,
we can calculate gap $r_{s}(\mathcal{N}_{st})$ in Eq.~(\ref{eq:r}) for the training tasks.

The probability of the evaluated data points is factorized using
a policy function:
\begin{align}
  p(\mathcal{N}_{st})=\prod_{\tau=1}^{t}p(n_{s\tau}|a(\cdot;\mathcal{N}_{s,\tau-1},\bm{\Psi})),  
\end{align}
where 
\begin{align}
  p(n_{st}|a(\cdot;\mathcal{N}_{s,t-1},\bm{\Psi}))
 =\left\{
  \begin{array}{ll}
    1 & \text{if $n_{st}=\argmax_{n\notin\mathcal{N}_{st}}a(\vec{x}_{sn};\mathcal{N}_{s,t-1},\bm{\Psi})$},\\
    0 & \text{otherwise},
  \end{array}
  \right.
  \label{eq:pn}
\end{align}
using Eq.~(\ref{eq:n}).
The data point with the highest policy function value is included in the evaluated data points.
We use the initially given evaluated data points for $\mathcal{N}_{s0}$.
This BO formulation can be regarded as a Markov decision process,
where evaluated data points $\mathcal{N}_{st}$ is the state,
data index to be evaluated the next $n_{s,t+1}$ is the action,
and gap $r_{s}(\mathcal{N}_{st})$ is the negative reward.
Therefore, we can train parameters $\bm{\Psi}$ of the policy function by
a reinforcement learning algorithm
described in Section~\ref{sec:train}.

\subsection{Policy function}
\label{sec:model}

\begin{figure*}[t!]
  \centering
  \includegraphics[width=33em]{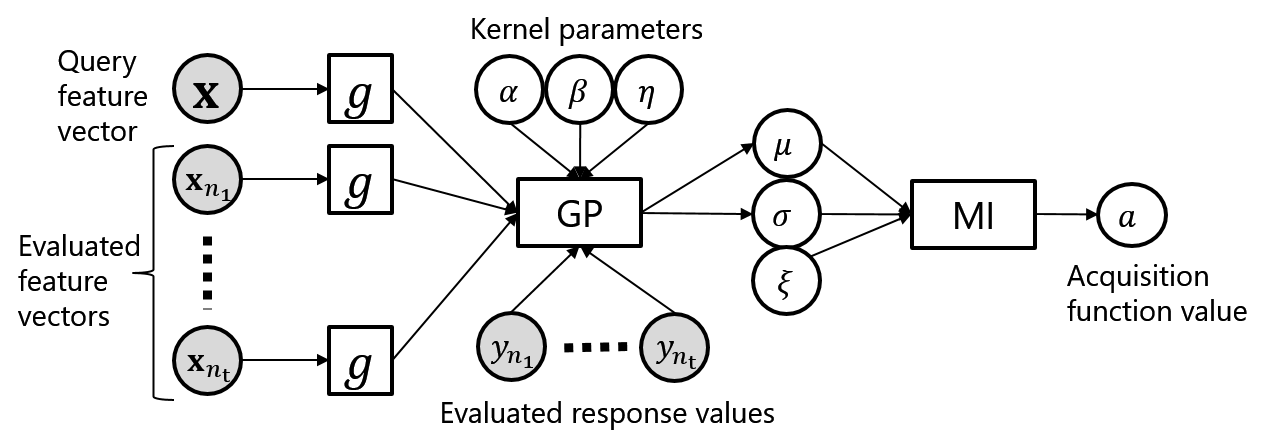}
  \caption{Our policy model takes query feature vector $\vec{x}$ and set of evaluated data points $\{(\vec{x}_{n_{\tau}},y_{n_{\tau}})\}_{\tau=1}^{t}$ as input and outputs acquisition function value $a$ of the query. First, query and observed feature vectors are nonlinearly transformed by neural network $g$. Second, mean $\mu$ and variance $\sigma^{2}$ of the response value at the query are estimated by GP using the transformed feature vectors, observed response values $\{y_{n_{\tau}}\}_{\tau=1}^{t}$, and kernel parameters $\alpha$, $\beta$, and $\eta$ in Eqs.~(\ref{eq:mu},\ref{eq:sigma}). Finally, acquisition function value $a$ of the query is calculated by mean $\mu$, variance $\sigma^{2}$, and $\xi$ based on GP-MI in Eq.~(\ref{eq:a}). We omit dependency between $\xi$ and past variance $\sigma^{2}$. White and gray circles represent observed and hidden variables. Rectangles represent functions.}
  \label{fig:model}
\end{figure*}

We model the policy function in Eq.~(\ref{eq:n})
by incorporating deep kernel learning (DKL)~\cite{wilson2016deep},
Gaussian process (GP), and
a mutual information-based acquisition function (MI)~\cite{contal2014gaussian}
as shown in Figure~\ref{fig:model}.
By using DKL,
we can flexibly learn representations from the given data,
which is especially important for such high-dimensional data
as text documents.
We share neural networks in deep kernels across different tasks,
enabling us to learn shared knowledge across tasks,
and use it for unseen test tasks.
MI has been successfully used for BO tasks.
By using MI as a component in our model,
we can exploit its useful knowledge designed for BO.
Instead of MI,
we can incorporate other types of acquisition functions designed for BO
if they are differentiable in the proposed method.

With MI, the acquisition function is given by
\begin{align}
  a(\vec{x};\mathcal{N}_{st},\bm{\Psi})=\mu(\vec{x};\mathcal{N}_{st},\bm{\Psi})
  +\sqrt{\nu}\left(\sqrt{\sigma^{2}(\vec{x};\mathcal{N}_{st},\bm{\Psi})+\xi_{st}}-\sqrt{\xi_{st}}\right),
  \label{eq:a}
\end{align}
where
$\mu(\vec{x};\mathcal{N}_{st},\bm{\Psi})$ and
$\sigma^{2}(\vec{x};\mathcal{N}_{st},\bm{\Psi})$ are
the mean and variance at $\vec{x}$ given $\mathcal{N}_{st}$,
\begin{align}
  \xi_{s,t+1}=\xi_{st}+\sigma^{2}(\vec{x};\mathcal{N}_{st},\bm{\Psi}),
\end{align}
and $\nu>0$ is a hyperparameter.
With a GP with a zero mean and kernel function $k(\cdot,\cdot;\bm{\Psi})$,
the mean and variance are given by
\begin{align}
  \mu(\vec{x};\mathcal{N}_{st},\bm{\Psi})=\vec{k}_{st}^{\top}(\vec{K}_{st}+\beta\vec{I})^{-1}\vec{y}_{st},
  \label{eq:mu}
\end{align}
\begin{align}
  \sigma^{2}(\vec{x};\mathcal{N}_{st},\bm{\Psi})=k(\vec{x},\vec{x};\bm{\Psi})-\vec{k}_{st}^{\top}\vec{K}_{st}^{-1}\vec{k}_{st},
  \label{eq:sigma}  
\end{align}
where $\vec{k}_{st}=[k(\vec{x},\vec{x}';\bm{\Psi})]_{\vec{x}'\in\mathcal{N}_{st}}\in\mathbb{R}^{|\mathcal{N}_{st}|}$
is a kernel vector between $\vec{x}$ and the evaluated data points at timestep $t$,
$\vec{K}_{st}=[k(\vec{x},\vec{x}';\bm{\Psi})]_{\vec{x},\vec{x}'\in\mathcal{N}_{st}}$
is the kernel matrix of the evaluated data points with size $(|\mathcal{N}_{st}| \times |\mathcal{N}_{st}|)$,
and $\beta>0$ is the observation noise parameter.

We use the RBF kernel with neural network transformation for DKL,
\begin{align}
  k(\vec{x},\vec{x}';\bm{\Psi})&=\alpha\exp\left(-\frac{1}{2\eta}\parallel g(\vec{x};\bm{\Theta})-g(\vec{x}';\bm{\Theta})\parallel^{2}\right)
  +\beta\delta(\vec{x},\vec{x}'),
  \label{eq:k}
\end{align}
where $g(\cdot;\bm{\Theta})$ is a neural network with parameter $\bm{\Theta}$,
$\alpha$ and $\eta$ are kernel parameters,
and $\delta$ is the Kronecker delta function,
i.e., $\delta(\vec{x},\vec{x}')=1$ if $\vec{x}=\vec{x}'$ and $\delta=0$ otherwise.
The parameters in the policy function
consist of the parameters of the neural networks and the kernel parameters,
$\bm{\Psi}=\{\bm{\Theta},\alpha,\beta,\eta\}$.
The policy function in Eq.~(\ref{eq:a})
acts as a neural network that
takes query point $\vec{x}$ and set of evaluated data points $\mathcal{N}_{st}$ as input,
and outputs a scalar value,
where neural network $g$ and GP are used as its components.
Since the number of evaluated data points $\mathcal{N}_{st}$ is different across
different timesteps,
standard neural networks cannot be used as a 
policy function when they assume that the input layer's size is fixed.
On the other hand, our model can handle different numbers of evaluated data points
using the GP framework, where the mean and variance given evaluated data points
are calculated in a closed form.

\subsection{Training procedure}
\label{sec:train}

We train parameters $\bm{\Psi}$ 
by minimizing the following expected cumulative gap with discount factor $\gamma\in(0,1]$,
\begin{align}
  L(\bm{\Psi}) = \mathbb{E}_{s}\left[\mathbb{E}_{a}\left[\sum_{t=1}^{T}\gamma^{t-1}r_{s}(\mathcal{N}_{st})\right]\right],
  \label{eq:L}
\end{align}
where evaluated data points $\mathcal{N}_{st}$ are selected
according to policy $a$ with parameter $\bm{\Psi}$.
Objective function $L(\bm{\Psi})$ is minimized by the policy gradient method~\cite{sutton1999policy} using the following gradient,
\begin{align}
  \frac{\partial L(\bm{\Psi})}{\partial\bm{\Psi}}=\mathbb{E}_{s}\left[\mathbb{E}_{a}\left[\sum_{t=1}^{T}G_{t}\frac{\partial\log p(n_{st}|a(\cdot;\mathcal{N}_{s,t-1},\bm{\Psi}))}{\partial\bm{\Psi}}
      \right]\right],
\end{align}
where $G_{t}=\sum_{\tau=t}^{T}\gamma^{\tau-t}r_{s\tau}$
is the sum of the future gaps with discount factor $\gamma$ at the $t$th timestep.
The training procedure
is shown in Algorithm~\ref{alg}.
For each training epoch, we first randomly sample a task in Line 3
and randomly generate initial evaluated data points
$\mathcal{N}_{s0}\subset\{1,\cdots,N_{s}\}$ in Line 4.
Next we generate
BO episode $n_{s1}, \mathcal{N}_{s1}, R_{s1}, \cdots, n_{sT}, \mathcal{N}_{sT}, R_{sT}$ using policy $a(\cdot;\mathcal{N}_{s,t-1},\bm{\Psi})$ and transition $\tilde{p}(n_{st}|a(\cdot;\mathcal{N}_{s,t-1},\bm{\Psi}))$
in Lines 5--10,
where we use the following stochastic transition function
\begin{align}
  \tilde{p}(n_{st}=n|a(\cdot;\mathcal{N}_{s,t-1},\bm{\Psi}))
  =\left\{
  \begin{array}{ll}
    \frac{\exp(a(\vec{x}_{sn};\mathcal{N}_{s,t-1},\bm{\Psi}))}
      {\sum_{n'\notin\mathcal{N}_{s,t-1}}\exp(a(\vec{x}_{sn'};\mathcal{N}_{s,t-1},\bm{\Psi}))}
     & \text{if $n_{s\tau}\notin\mathcal{N}_{s,t-1}$}\\
    0 & \text{otherwise},
  \end{array}
  \right.      
\label{eq:pntilde}
\end{align}
instead of the deterministic transition in Eq.~(\ref{eq:pn}).
Next we calculate the sum of future gaps $G_{t}$ in Lines 11--13.
Then, we calculate loss
\begin{align}
  \tilde{L}=\sum_{t=1}^{T}G_{t}\log\tilde{p}(n_{st}|a(\cdot|\mathcal{N}_{s,t-1},\bm{\Psi})),
  \label{eq:Ltilde}
\end{align}
and update parameters $\bm{\Psi}$ by minimizing the loss using a stochastic gradient method in Lines 14--15.
Since loss $L$ is differentiable, we can backpropagate it
through the MI to update the neural network parameters 
and the kernel parameters.
We use the average of the discounted sum of the future rewards
as a baseline to reduce the
variance~\cite{weaver2001optimal,greensmith2004variance}.

\begin{algorithm}[t!]
  \caption{Training procedure of our model.}
  \label{alg}
  \begin{algorithmic}[1]
    \renewcommand{\algorithmicrequire}{\textbf{Input:}}
    \renewcommand{\algorithmicensure}{\textbf{Output:}}
    \REQUIRE{Datasets $\{\{(\vec{x}_{sn},y_{sn})\}_{n=1}^{N_{s}}\}_{s=1}^{S}$,
      number of queries $T$, discount factor $\gamma$}
    \ENSURE{Trained model parameters $\bm{\Psi}$}
    \STATE Initialize model parameters $\bm{\Psi}$.
    \WHILE{End condition is satisfied}
    \STATE Randomly sample a task, $s\sim\mathrm{Uniform}(1,\cdots,S)$.
    \STATE Randomly sample initial evaluated data points $\mathcal{N}_{s0}$.
    \FOR{$t \in \{1,\cdots,T\}$}    
    \STATE Calculate acqusition function $a(\vec{x};\mathcal{N}_{s,t-1},\bm{\Psi})$ by Eq.~(\ref{eq:a}) for $\vec{x}\in\{\vec{x}_{sn}\}_{n\notin\mathcal{N}_{s,t-1}}$.
    \STATE Sample query data index $n_{st}$ according to Eq.~(\ref{eq:pntilde}).
    \STATE Update evaluated data points $\mathcal{N}_{st}\gets\mathcal{N}_{s,t-1}\cup n_{sn}$.
    \STATE Calculate negative reward $r_{s}(\mathcal{N}_{st})$.
    \ENDFOR
    \FOR{$t \in \{1,\cdots,T\}$}
    \STATE Calculate discounted sum of future negative rewards
    $G_{t}=\sum_{\tau=t}^{T}\gamma^{\tau-t}r_{s}(\mathcal{N}_{s\tau})$.
    \ENDFOR
    \STATE Calculate loss $\tilde{L}$ in Eq.~(\ref{eq:Ltilde}) and its gradient.
    \STATE Update model parameters $\bm{\Psi}$ using loss $L$ and its gradient using a stochastic gradient method.
    \ENDWHILE
  \end{algorithmic}
\end{algorithm}

\section{Experiments}
\label{sec:experiments}

\subsection{Data}

We experimentally evaluated the effectiveness of our proposed method
using text document datasets
with a BO task to find a document that resembles
an unknown target document with fewer evaluations.
This BO task simulated a recommendation system, where
a user seeks a document that matches her preferences
by evaluating a limited number of documents.
The user's evaluation for each document is the target function,
where evaluation is expensive since the documents must be read.
We represented a document as bag-of-words $\vec{x}_{sn}=(x_{sn1},\cdots,x_{snJ})$, or a word frequency vector,
where $x_{snj}$ is the number of occurrences of the $j$th vocabulary term in the $n$th document for the $s$th user,
and $J$ is the vocabulary size.
Each user was a task, where preferences were different across users,
but the target functions of different users were related since the vocabulary terms were shared across users,
and some users have similar preferences.

We used the following three text document datasets: Reuter, NeurIPS,
and WebKB.
The Reuter data were constructed from documents that appeared on the Reuters newswire categorized into eight classes~\footnote{The original Reuter data were obtained from ~\url{http://www.daviddlewis.com/resources/testcollections/reuters21578/}.}~\cite{apte1994automated}.
The NeurIPS data were constructed from conference papers
in Neural Information Processing Systems 1--12
categorized into 13 research topics~\footnote{The original NeurIPS data were obtained from~\url{https://cs.nyu.edu/~roweis/data.html}.}.
The WebKB data were constructed from web pages categorized into four universities~\footnote{The original WebKB data were obtained from~\url{http://www.cs.cmu.edu/afs/cs.cmu.edu/project/theo-20/www/data/}.}~\cite{cravenlearning}.
We omit vocabulary terms that occurred in less than 50 documents
as well as documents that contained fewer than 50 words.
The number of documents was 1,909 and the number of vocabulary terms was 799
in the 20News data,
592 and 1,655 in the NeurIPS data, and 3,264 and 1,365 in the WebKB data.

We calculated the similarity
by the negative Euclidean distance
between the representations obtained by neural networks.
For the neural network,
we used a four-layered feed-forward neural network with 32 hidden units
and a rectified linear unit, $\mathrm{ReLU}(x)=\max(0,x)$, for the activation.
The neural network was trained by minimizing the cross-entropy loss using category labels
as in the text classification.
We used the last hidden layer as the representations of documents.
By transforming the word frequency vectors to the representations by the neural network,
the similarity reflected the topics of documents that were related to the category labels.
For each BO task,
we randomly selected a target document and 500 candidate documents,
and calculated their similarity,
where the target document was considered the user's most preferred document.
We generated $\{20,40,\cdots,100\}$ training, 20 validation, and 50 test tasks
for each dataset.
For each validation and test task,
an initial evaluated data point was randomly selected.
We ran ten experiments for each dataset,
and evaluated the performance based on the average of the cumulative gaps,
$\hat{R} = \frac{1}{ST}\sum_{s=1}^{S}\sum_{t=1}^{T}r_{s}(\mathcal{N}_{st})$,
in the test BO tasks.

\subsection{Proposed method setting}

We used a four-layered feed-forward neural network
with 32 hidden units and ReLU activation
for neural network $g$ in the kernel functions.
The discount factor was $\gamma=0.99$.
The number of queries was 10 for each task,
and the number of initial evaluated data points was $|\mathcal{N}_{s0}|=1$.
We used three types of acquisition functions: MI,
expected improvement (EI)~\cite{mockus1978application,jones2001taxonomy},
and upper confidence bound (UCB)~\cite{auer2002using,srinivas2010gaussian}.
With EI,
the acquisition function is given by
\begin{align}
  a(\vec{x};\mathcal{N}_{st},\bm{\Psi})&=
  \left(\mu(\vec{x};\mathcal{N}_{st},\bm{\Psi})-\max_{n\in\mathcal{N}_{st}}y_{sn}\right)\Phi(\vec{x}')
  \nonumber\\
  &+\sigma(\vec{x};\mathcal{N}_{st},\bm{\Psi})\phi(\vec{x}'),
  \label{eq:a_ei}
\end{align} 
where $\phi$ and $\Phi$ are the probability and cumulative density functions of the standard normal distribution,
and
\begin{align}
  \vec{x}'=\frac{\mu(\vec{x};\mathcal{N}_{st},\bm{\Psi})-\max_{n\in\mathcal{N}_{st}}y_{sn}}{\sigma(\vec{x};\mathcal{N}_{st},\bm{\Psi})}.
\end{align}
With UCB, 
the acquisition function is given by
\begin{align}
  a(\vec{x};\mathcal{N}_{st},\bm{\Psi})=\mu(\vec{x};\mathcal{N}_{st},\bm{\Psi})
  +\sqrt{\nu}\sigma(\vec{x};\mathcal{N}_{st},\bm{\Psi}),
  \label{eq:a_ucb}
\end{align}
where $\nu>0$ is the hyperparameter.
For the hyperparameter of MI and UCB, we used $\nu=\frac{\log 2}{10^{-6}}$ like in a previous work~\cite{contal2014gaussian}.
We optimized the model parameters using Adam~\cite{kingma2014adam} with learning rate $10^{-3}$
and batch size 16.
The validation data were used for early stopping, and the maximum number of epochs was 1,000.
We pretrained the model parameters by maximizing the marginal likelihood.
We implemented the proposed method with PyTorch~\cite{paszke2017automatic}.

\subsection{Comparing methods}

We compared the proposed method with
the following five methods: DKL, GP, RL, MetaBO, and Random.
DKL is deep kernel learning~\cite{wilson2016deep},
where GPs with neural network-based kernels
were trained by maximizing the marginal likelihood,
\begin{align}
  \sum_{s=1}^{S}\sum_{n=1}^{N_{s}}\log\rm{Normal}(\vec{y}_{s}|\vec{0},\vec{K}_{s}),
  \label{eq:marginal}
\end{align}
where $\vec{y}_{s}=(y_{s1},\cdots,y_{sN_{s}})$ is the vector of response values in the $s$th task,
$\vec{K}_{s}\in\mathbb{R}^{N_{s}\times N_{s}}$ is the kernel matrix between feature vectors in the $s$th task,
and $\rm{Normal}(\cdot|\bm{\mu},\bm{\Sigma})$ is the multivariate normal distribution
with mean $\bm{\mu}$ and covariance $\bm{\Sigma}$.
We used the same neural network architecture with the proposed method.
GP is a Gaussian process with an RBF kernel,
where kernel parameters $\alpha$, $\beta$, and $\eta$
were trained by maximizing the marginal likelihood in Eq.(\ref{eq:marginal}).
In DKL and GP, we used three types of acquisition functions: MI, EI, and UCB.

RL is a reinforcement learning-based method with a policy modeled by neural networks
as in a previous work~\cite{vishnu2019meta}.
For the neural networks, we used deep sets~\cite{zaheer2017deep}
for taking a set of evaluated data points as input,
\begin{align}
  \vec{z}_{st}=g_{\mathrm{RL}}\left(\frac{1}{|\mathcal{N}_{st}|}\sum_{n'\in\mathcal{N}_{st}}f_{\mathrm{RL}}(\vec{x}_{sn'})\right),
\end{align}
where $\vec{z}_{st}$ is a vector representation of the evaluated data points at the $t$th timestep,
and
$f_{\mathrm{RL}}$ and $g_{\mathrm{RL}}$ are three-layered feed-forward neural networks with 32 hidden units.
The policy was modeled by concatenating the representation and
a query data point,
\begin{align}
  p_{\mathrm{RL}}(n_{st}=n|\mathcal{N}_{s,t-1})
  \propto
  \left\{
  \begin{array}{ll}
    \exp(h_{\mathrm{RL}}([\vec{x}_{sn},\vec{z}_{s,t-1}]))
     & \text{if $n_{s\tau}\notin\mathcal{N}_{s,t-1}$}\\
    0 & \text{otherwise},
  \end{array}
  \right.      
\label{eq:pnrl}
\end{align}
where $[\cdot,\cdot]$ represents the concatenation,
and $h_{\mathrm{RL}}$ is a four-layered feed-forward neural network with 32 hidden units.

MetaBO is a meta-learning method for BO based on reinforcement learning~\cite{volpp2019meta},
where the acquisition functions were modeled by a neural network that took
the GP's outputs and query data point,
\begin{align}
  p_{\mathrm{M}}(n_{st}=n|\mathcal{N}_{s,t-1})
  \propto
  \left\{
  \begin{array}{ll}
    \exp(h_{\mathrm{M}}([\mu(\vec{x}_{sn}),\sigma^{2}(\vec{x}_{sn}),\vec{x}_{sn}]))
     & \text{if $n_{s\tau}\notin\mathcal{N}_{s,t-1}$}\\
    0 & \text{otherwise},
  \end{array}
  \right.      
\label{eq:pnmb}
\end{align}
where $h_{\mathrm{M}}$ is a four-layered feed-forward neural network with 32 hidden units,
and $\mu(\vec{x}_{sn})$ and $\sigma^{2}(\vec{x}_{sn})$ are the mean and variance
estimated by the GP with RBF kernels.
The kernel parameters were trained by maximizing the marginal likelihood.
With RL and MetaBO, we trained neural networks
by the policy gradient method, as in our proposed method.

\subsection{Results}

\begin{figure*}[t!]
  \centering
  {\tabcolsep=0em
  \begin{tabular}{ccc}
    & Reuter &\\
    \includegraphics[height=9.8em]{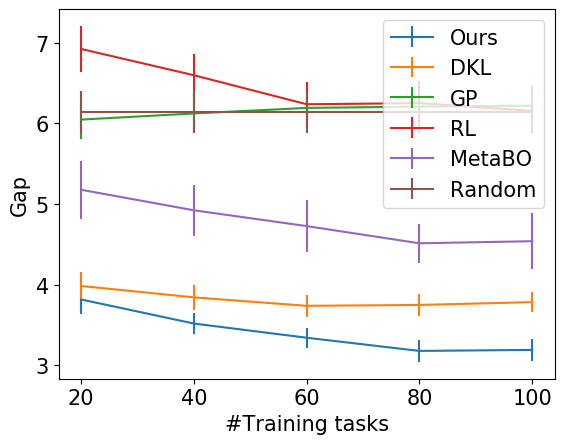}&
    \includegraphics[height=9.8em]{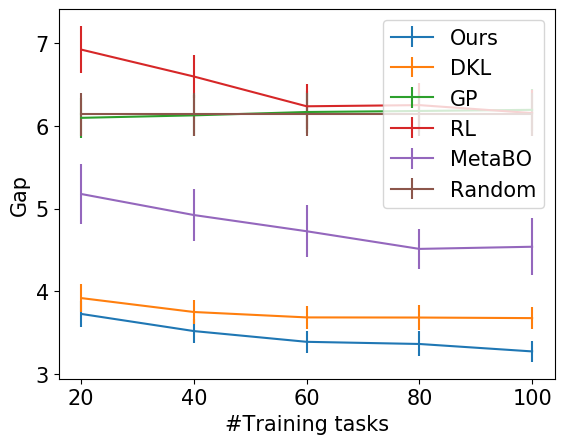}&
    \includegraphics[height=9.8em]{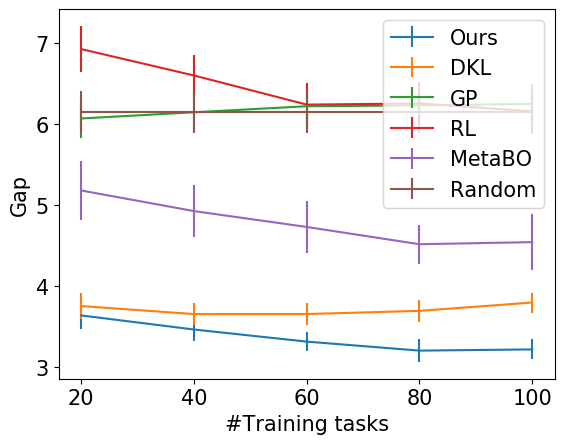}\\
    & NeurIPS &\\
    \includegraphics[height=9.8em]{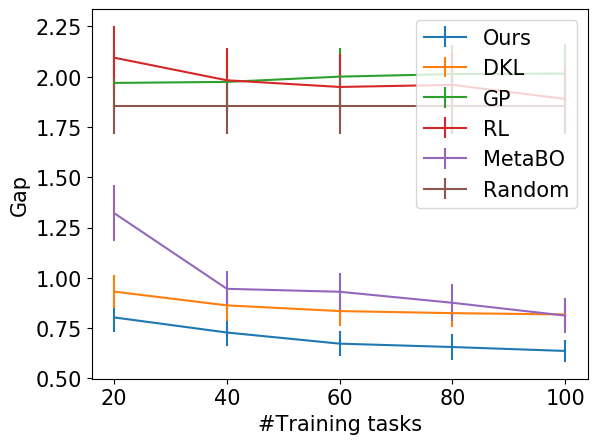}&
    \includegraphics[height=9.8em]{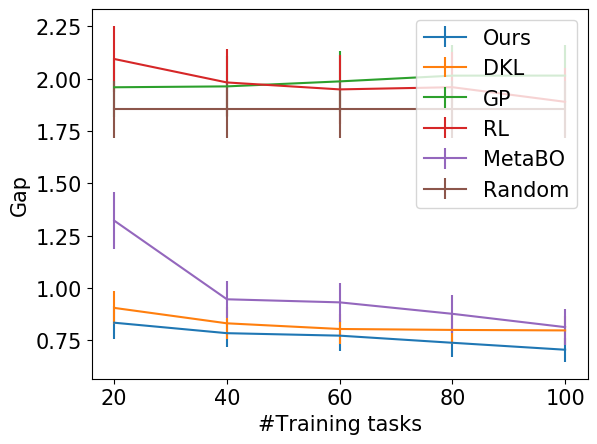}&
    \includegraphics[height=9.8em]{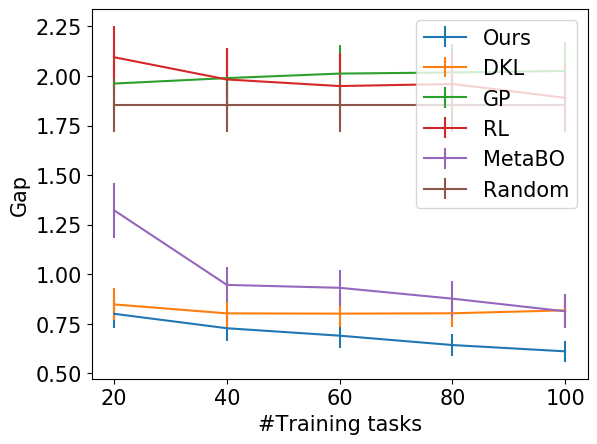}\\
    &WebKB&\\
    \includegraphics[height=9.8em]{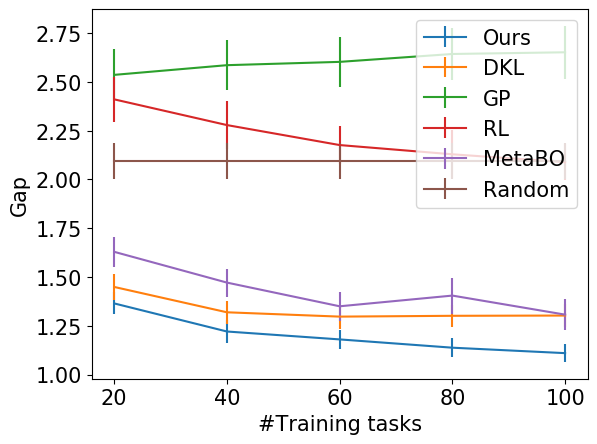}&
    \includegraphics[height=9.8em]{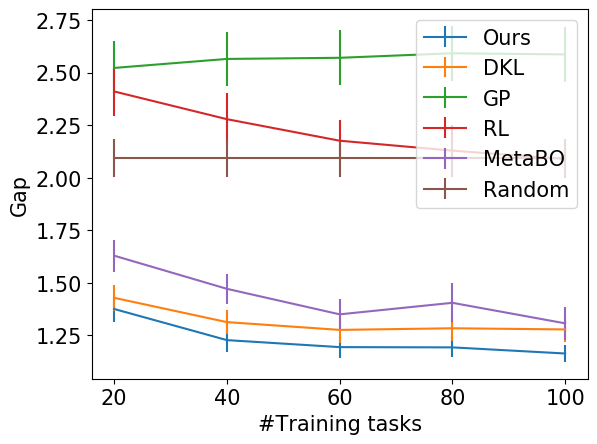}&
    \includegraphics[height=9.8em]{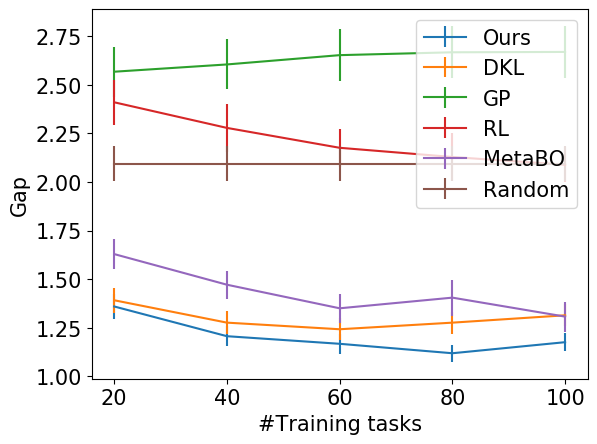}\\
    (a) MI & (b) EI & (c) UCB \\
  \end{tabular}}
  \caption{Average test cumulative gap with different numbers of training tasks on Reuter, NeurIPS, and WebKB data.
    Bars show standard error.
    We used three kinds of acquisition functions (a) MI, (b) EI, and (c) UCB with the proposed method, DKL,
    and GP.
    RL, MetaBO, and Random are the identical among (a), (b), and (c).}
  \label{fig:result_ndataset}
\end{figure*}

Figure~\ref{fig:result_ndataset} shows the average test cumulative gap
with different numbers of training tasks.
The proposed method achieved the lowest gap in all cases.
As the number of training tasks increased,
the gap decreased with the proposed method, RL, and MetaBO,
indicating that the reinforcement learning-based methods
learned how to select queries in BO using a wide variety of training tasks.
With DKL, the decrease based on the number of training tasks
was smaller than the reinforcement learning-based methods
because it was trained by fitting to the training data by maximizing the marginal likelihood.
Thus it did not directly improve the BO performance.
RL's gap was high, especially when the number of training tasks was small.
Since RL needs to train the policy from scratch using training tasks,
it requires many training tasks.
On the other hand, since MetaBO used GP's result for modeling the policy,
it outperformed RL.
Furthermore, for modeling the policy, the proposed method uses
acquisition functions, i.e., MI, EI, or UCB, which were
developed for BO by experts.
Therefore, the proposed method outperformed MetaBO and RL.
GP's gap was high,
indicating that learning kernels with neural networks
is important for modeling high-dimensional data.

\begin{figure*}[t!]
  \centering
  \begin{tabular}{ccc}
    \includegraphics[height=9.3em]{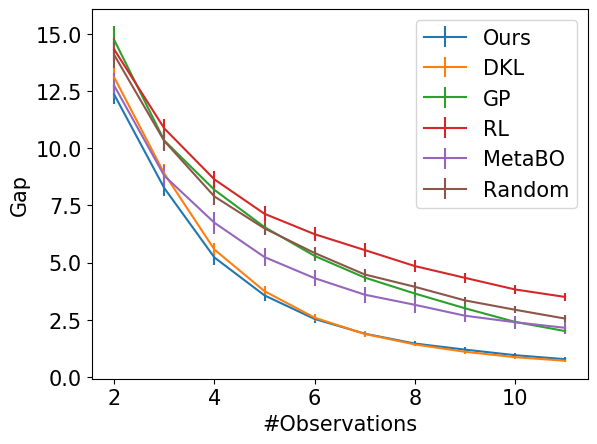}&
    \includegraphics[height=9.3em]{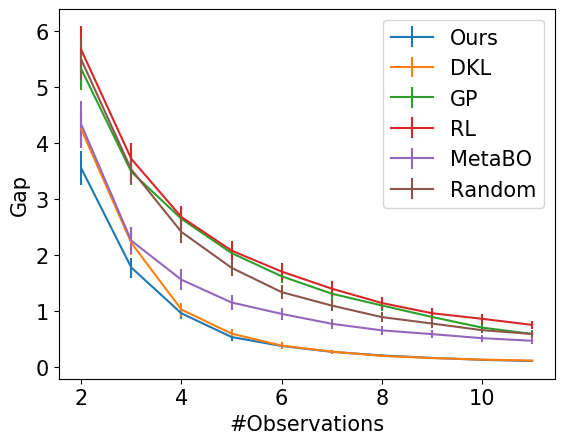}&
    \includegraphics[height=9.3em]{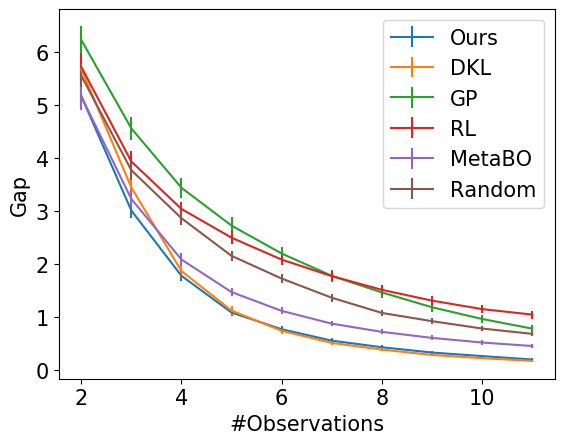}\\
    (a) Reuter & (b) NeurIPS & (c) WebKB \\
  \end{tabular}
  \caption{Average test gap with different numbers of evaluated data points on Reuter, NeurIPS, and WebKB data.
    We used MI acquisition functions, and 20 training tasks.
    Bars show standard error.}
  \label{fig:result_observation}
\end{figure*}

Figure~\ref{fig:result_observation}
shows the average test gap with different amounts of evaluated data points using MI acquisition functions.
The gaps of the proposed method and DKL were almost the same when the number of the evaluated data points was large.
However, the gap of the proposed method was better than DKL when the number of the evaluated data points was small.
This result demonstrates the effectiveness of training neural networks by minimizing the expected cumulative gap
for finding higher response values with fewer evaluations.
Although MetaBO's gap was low when the number of evaluated data points was relatively small,
the gap's decrease based o the number of evaluated data points was slower than the proposed method.

\begin{table}[t!]
  \centering
  \caption{Training time in seconds.}
  \label{tab:time}
  \begin{tabular}{lrrrrr}
    \hline
    & Reuter & NeurIPS & WebKB \\
    \hline
    Ours & 931.886 & 1080.168 & 993.435 \\
    DKL & 131.102 & 224.183 & 201.535 \\
    GP & 150.889 & 251.762 & 215.716 \\
    RL & 451.380 & 637.127 & 583.239 \\
    MetaBO & 934.847 & 1079.129 & 1064.673 \\
    \hline
  \end{tabular}
\end{table}

Table~\ref{tab:time}
shows the computational time in seconds for training with computers with GeForce GTX 1080 Ti GPUs.
The training times with DKL and GP, which were trained with the supervised learning framework,
were shorter than the other methods, which were trained with a reinforcement learning framework.
It is because reinforcement learning-based methods run BO for each training epoch.
The proposed method and MetaBO required more training time than RL
since they calculated GPs in their methods.

\section{Conclusion}
\label{sec:conclusion}

We proposed a meta-learning method
for learning neural network-based kernels for BO
to minimize the expected cumulative gap between the true optimum value
and the best found value.
The proposed method learns knowledge in various training tasks,
and can use it for unseen test tasks by sharing neural networks.
Since the proposed method incorporates acquisition functions,
such as mutual information and expected improvement, 
for modeling a policy that selects the next data point to be evaluated,
we can employ the knowledge in the existing acquisition functions,
which enables us to learn policies with fewer training tasks.
Although our results are encouraging,
we must extend our approach in several future directions.
First, 
we will apply it to other types of high-dimensional structured data,
such as images and time-series,
using convolutional and recurrent neural networks in kernel functions.
Second, we plan to improve our proposed method using
reinforcement learning techniques, which includes actor-critic methods~\cite{grondman2012survey}.

\bibliographystyle{abbrv}
\bibliography{arxiv_meta_bo}

\end{document}